\let\oldvec\vec
\documentclass{llncs}
\let\vec\oldvec
\usepackage{graphicx}
\usepackage{epsfig}
\usepackage[T1]{fontenc}
\usepackage[utf8]{inputenc}
\usepackage{tikz}
\usetikzlibrary{arrows,positioning,shapes}
\usepackage{microtype}
\usepackage{epstopdf}
\usepackage{subcaption}
\usepackage{wrapfig}
\usepackage{underscore}

\begin{document}

\pagestyle{empty}
\mainmatter

\title{On the Effectiveness of Genetic Operations in Symbolic Regression}

\author{Bogdan Burlacu\inst{1,2} \and Michael Affenzeller\inst{1,2} \and Michael Kommenda\inst{1,2}}

\institute {
    Heuristic and Evolutionary Algorithms Laboratory\\
    School of Informatics, Communications and Media\\
    University of Applied Sciences Upper Austria\\
    Softwarepark 11, 4232 Hagenberg, Austria\\
 \vspace{0.2cm}
    \and Institute for Formal Models and Verification\\
    Johannes Kepler University  Linz\\
    Altenbergerstr. 69, 4040 Linz, Austria\\
 \vspace{0.2cm}
	\email{\{bogdan.burlacu,michael.affenzeller,michael.kommenda\}@fh-hagenberg.at}\\
    \thanks{The final publication is available at\\ \url{https://link.springer.com/chapter/10.1007/978-3-319-27340-2_46}}    
}

\maketitle

\begin{abstract}
	This paper describes a methodology for analyzing the evolutionary dynamics of genetic programming (GP) using genealogical information, diversity measures and information about the fitness variation from parent to offspring. We introduce a new subtree tracing approach for identifying the origins of genes in the structure of individuals, and we show that only a small fraction of ancestor individuals are responsible for the evolvement of the best solutions in the population.
\end{abstract}


\section{Introduction}   
\label{sec:introduction}
Empirical analysis in the context of different benchmark problems and tentative algorithmic improvements (such as various selection schemes or fitness assignment techniques) has a limited ability of explaining genetic programming (GP) behavior and dynamics. Results usually confirm our intuitions about the relationship between selection pressure, diversity, fitness landscapes and genetic operators, but they prove difficult to extend to more general theories about the internal functioning of GP.  

This work is motivated by the necessity for a different approach to study the GP evolutionary process. Instead of looking for correlations between different selection or fitness assignment mechanisms and solution quality or diversity, we focus on the reproduction process itself and the effectiveness of the variation-producing operators in transferring genetic material.

Achieving good solutions depends on the efficient use of the available gene pool given its inherent stochasticity (random initialization, random crossover, random mutation). Under the effects of selection pressure, many suboptimal exchanges of genetic information will cause a decrease in the amount of genetic material available to the evolutionary engine. Measures to mitigate this phenomenon usually use various heuristics for guiding either selection or the crossover operator towards more promising regions of the search space \cite{gustafson:2003:iidigpbaaoteop,gustafson:2004:IEEE}.     

Diversity is an important aspect of GP, considered to be a key factor in its performance. Multiple studies dedicated to GP diversity analyze diversity measures (based on various distance metrics, for example \cite{Mattiussi2004-ID509}) in correlation with the effects of genetic operators \cite{ekart:2002:EuroGP,Nguyen:2006:ASPGP,Jackson:2010:EuroGP}.
Genotype operations -- crossover in particular -- often have a negative (or at most, neutral) effect on individuals, leading to diversity loss in the population following each selection step. This effect is due to the interplay between crossover and selection which leads to an increase in average program size \cite{Dignum:2008:eurogp2} (when sampling larger programs, crossover has a higher chance of having a neutral effect).

\section{Methodology}
In this paper we introduce a new methodology for the exact identification (``tracing'') of any structural change an individual is subjected to during evolution. We use this methodology in combination with population diversity and genealogy analysis methods to investigate the effects of the genetic operators in terms of how often they lead to a fitness improvement, how often they overlap (for example when the same area inside the tree is repeatedly targeted by crossover), and how often they cancel each other out. 

\subsection{Tracing of genotype fragments}
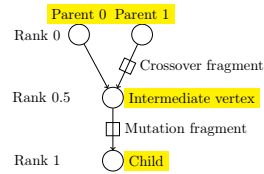
\begin{wrapfigure}{r}{0.3\textwidth}
	\vspace{-25pt}
	\resizebox{0.3\textwidth}{!}{
		\begin{tikzpicture}
	\tikzset{vertex/.style={draw,inner sep=5pt,circle}}
	\tikzset{fragment/.style={draw,diamond,inner sep=3pt}}
	\node[vertex,label={[fill=yellow]90:Parent 0}] (1) at (0,0) {};
	\node[vertex,label={[fill=yellow]90:Parent 1},right=of 1] (2) {};
	\node[vertex,label={[fill=yellow]0:Intermediate vertex},below=of 2,xshift=-0.7cm] (3) {};
	\node[vertex,label={[fill=yellow]0:Child},below=of 3] (4) {};

	\node[left of=1] (r0) {Rank 0}; 
	\node[below of=r0,yshift=-0.5cm,xshift=0.1cm] (r1) {Rank 0.5}; 
	\node[below of=r1,yshift=-0.5cm,xshift=-.1cm] (r2) {Rank 1}; 

	\draw[->] (1) -- (3) ;
	\draw[->] (2) -- (3) node[fragment,midway,rotate=20,label={[distance=10pt]-20:Crossover fragment}]{};
	\draw[->] (3) -- (4) node[draw,rectangle,inner sep=4.5pt,midway,label={[distance=10pt]0:Mutation fragment}]{};

\end{tikzpicture}
	}
	\caption{Saving intermediate results in the genealogy graph}\label{fig:intermediate-vertex}
\end{wrapfigure}

This method is based on previous work on population genealogies \cite{Burlacu:2012:EMSS,Burlacu:2013:GECCOcomp}. During the algorithm run, every new generation is added to the genealogy graph with arcs connecting child vertices to their parents. 
When crossover is followed by mutation, both the results of crossover and mutation are saved in the graph (Figure \ref{fig:intermediate-vertex}).  

We define an individual's \textit{trace graph} as a collection of vertices representing its ancestors from which the various parts of its genotype originated, connected by a collection of arcs representing the different genotype operations that gradually assembled those parts. 

\begin{figure}[ht]
	\centering
	\resizebox{0.7\textwidth}{!}{
		\begin{tikzpicture}	
	\tikzset{
		v/.style={draw,inner sep=2pt,circle,minimum size=15pt},
		a/.style={draw,inner sep=2pt,circle,minimum size=15pt,fill=blue!70!white!20},
		b/.style={draw,inner sep=2pt,circle,minimum size=15pt,fill=red!30!white},
		c/.style={draw,inner sep=2pt,circle,minimum size=15pt,fill=green!30!white}
	}
	\begin{scope}[level distance=10mm,sibling distance=8mm]
		\node[a,label=-90:0,label=90:{Root parent}] (a) {$\times$}
		child { node[a,xshift=-0.5cm,label=-90:1] {$+$} 
			child { node[a,label=-90:2] {$a$} }	
			child { node[a,label=-90:3] {$b$} }	
		}
		child { node[b,xshift=0.5cm,label=-90:4] (swapped-subtree) {$-$} 
			child { node[b,label=-90:5] {$a$} }	
			child { node[b,label=-90:6] {$b$} }	
		};
		\node[v,label=-90:0,label=90:{Non-root parent},right of=a,xshift=2.7cm] (b) {$-$}
		child { node[c,xshift=-0.5cm,label=-90:1] (fragment) {$\times$} 
			child { node[c,label=-90:2] {$a$} }	
			child { node[c,label=-90:3] {$a$} }	
		}
		child { node[v,xshift=0.5cm,label=-90:4] {$\times$} 
			child { node[v,label=-90:5] {$b$} }	
			child { node[v,label=-90:6] {$b$} }	
		};
		\node[a,label=-90:0,label=90:{Child}, right of=b,xshift=2.7cm,label={[align=left,xshift=-2mm,yshift=-4mm]180:Inherited\\ part}] (c) {$\times$}
		child { node[a,xshift=-0.5cm,label=-90:1] {$+$} 
			child { node[a,label=-90:2] {$a$} }	
			child { node[a,label=-90:3] {$b$} }	
		}
		child { node[c,xshift=0.5cm,label=-90:4,label={[align=left]0:Received\\ fragment}] {$\times$} 
			child { node[c,label=-90:5] {$a$} }	
			child { node[c,label=-90:6] {$a$} }	
		};
	\end{scope}

	\node[above=of swapped-subtree,yshift=-12mm](m) {};
	\node[above=of fragment,yshift=-12mm](n) {};

	\path (m.north) edge[dashed,<-,bend left=60] node[text centered,yshift=2mm] {Subtree swap} (n.north) {};

\end{tikzpicture}
	}
	\caption{Preorder arithmetics for subtree inclusion}\label{fig:preorder-arithmetics}
\end{figure}
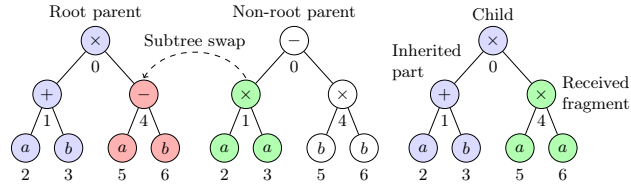

The tracing procedure uses a set of simple arithmetic rules to navigate genealogies and identify the relevant subtrees, based on the indices of the subtree to be traced and the index of the received fragment  (Figure \ref{fig:preorder-arithmetics}).  The nodes in each tree are numbered according to their preorder index $i$ such that, given two subtrees $A$ and $B$, $B \subset A$ if $i_A < i_B < i_A + l_A$, where $i_A$, $i_B$ are their respective preorder indices and $l_A$, $l_B$ are their lengths. 

Since some individuals within the ancestry of the traced individual may have contributed parts of their genotype to multiple offspring, there may exist multiple evolutionary trajectories in the trace graph passing through the same vertex or sequence of vertices, reflected in the graph by multiple arcs between the same two vertices, each arc representing the transmission of different genes or building blocks.

\subsection{Analysis of population dynamics}
The various measurements used to quantify the behavioral aspects of GP are described in more detail within the following paragraphs.

\noindent
\textbf{Genetic Operator Effectiveness}\\
Operator effectiveness is calculated as the difference in fitness between the child and its root parent.
\paragraph{Average fitness improvement} Let $N$ be the total number of individuals in the population, $t_i$ one individual and $p_i$ its parent:
\[
	\bar{q} = \frac{1}{N} \cdot \sum_{i=1}^{N} \big( Fitness(t_i) - Fitness(p_i) \big)
\]
\paragraph{Best fitness improvement}
Return the difference between the fitness values of the best individual $t_{best}$ and its parent $p_{best}$
\[
	q_{best} = Fitness(t_{best}) - Fitness(p_{best}) 
\]
The average and best fitness improvements are calculated individually for crossover and mutation operations.

\noindent
\textbf{Average relative overlap}\\
We define the relative overlap between two sets $A_1$ and $A_2$ using the S{\o}rensen-Dice coefficient\footnote{It was also possible to use the \textit{Jaccard index} $J(A_1,A_2)=\frac{|A_1 \cap A_2|}{|A_1 \cup A_2|}$ as it is very similar to the S{\o}rensen-Dice coefficient. However this choice makes no practical difference for the results presented in this publication} which can also be seen as a similarity measure between sets:
\[
	s(A_1, A_2) = \frac{2 \cdot |A_1 \cap A_2|}{|A_1| + |A_2|}
\]
The reason for using this measure is to see how much overlap exists between the trace graphs and root lineages of the individuals in the population. A high relative overlap would mean that diversity is exhausted as all the individuals have the same parents or ancestors. 

\noindent
\textbf{Genotype and phenotype similarity}\\
These similarity measures provide information about the evolution of diversity from both a structural (genotype) and a semantic (phenotype) perspective. Genotype similarity is calculated using a \textit{bottom-up tree mapping} \cite{Valiente01anefficient} that can be computed in time linear in the size of the trees and has the advantage that it works equally well for unordered trees. For two trees $T_1$ and $T_2$ and a bottom-up mapping $M$ between them, the similarity is given by:
\[
	GenotypeSimilarity(T_1,T_2) =\frac{2 \cdot |M|}{|T_1|+|T_2|}
\]
Phenotype similarity between two trees is calculated as the Pearson $R^2$ correlation coefficient between their respective output values on the training data.

\noindent
\textbf{Contribution ratio}\\
While it is clear that under the influence of random evolutionary forces (such as genetic drift or hitchhiking) each of an individual's ancestors plays an equally important role in the events leading to its creation, the trace graph represents a powerful tool for analyzing the origin of genes and the way solutions are assembled by the genetic algorithm. 

The size of the trace graph relative to the size of the complete ancestry can be used as a measure of the effort spent by the algorithm to achieve useful adaptation. For example, a small trace graph means that a small number of an individual's ancestors contributed to the assembly of its genotype, via an equally small number of genetic operations (crossover and mutation). The effort, seen as the ratio of effective genetic operations over the total number of genetic operations, can give an indication of how easy new and better solutions can be assembled by the algorithm.

The contribution ratio $r$ is given by the percentage of individuals from the best solution ancestry that had an actual contribution to its structure:
\[
	r = \frac{|Trace(\mathrm{bestSolution})|}{|Ancestry(\mathrm{bestSolution})|}
\]

\section{Experiments}
For the experimental part, we use GP to solve two symbolic regression benchmark problems:\smallskip\noindent\\
\textbf{Vladislavleva-8}
\[
	F_8(x_1, x_2) = \frac{(x_1 - 3)^4 + (x_2 - 3)^3 - (x_2 - 3)}{(x_2 - 2)^4 + 10}
\]
\textbf{Poly-10}
\[
	F(\mathbf{x}) = x_1 x_2 + x_3 x_4 + x_5 x_6 + x_1 x_7 x_9 + x_3 x_6 x_{10}
\]

The Vladislavleva-8 problem was solved using the standard GP algorithm (SGP) with a population size of 500 individuals and 50 generations (in order to be able to compute the trace graphs of each individual in the population in feasible time). 
For the Poly-10 problem the offspring selection GP (OSGP) \cite{Affenzeller:GAGP} was also tested with a population size of 300 individuals and gender-specific selection.

We analyzed the algorithm dynamics using the genealogy graph, the ancestry of the best solution and the trace history of its genotype. Other additional measurements such as diversity, size and quality distributions were included for a more complete picture. All the results were averaged on a collection of 20 algorithmic runs for each problem and algorithm configuration.

\begin{figure}[ht]
	\graphicspath{{results/vladislavleva8/sgp/tournament-selector-size-3/}}
	\captionsetup[subfigure]{justification=centering}
	\begin{subfigure}[b]{0.5\textwidth}
		\resizebox{1\textwidth}{!}{
			\input{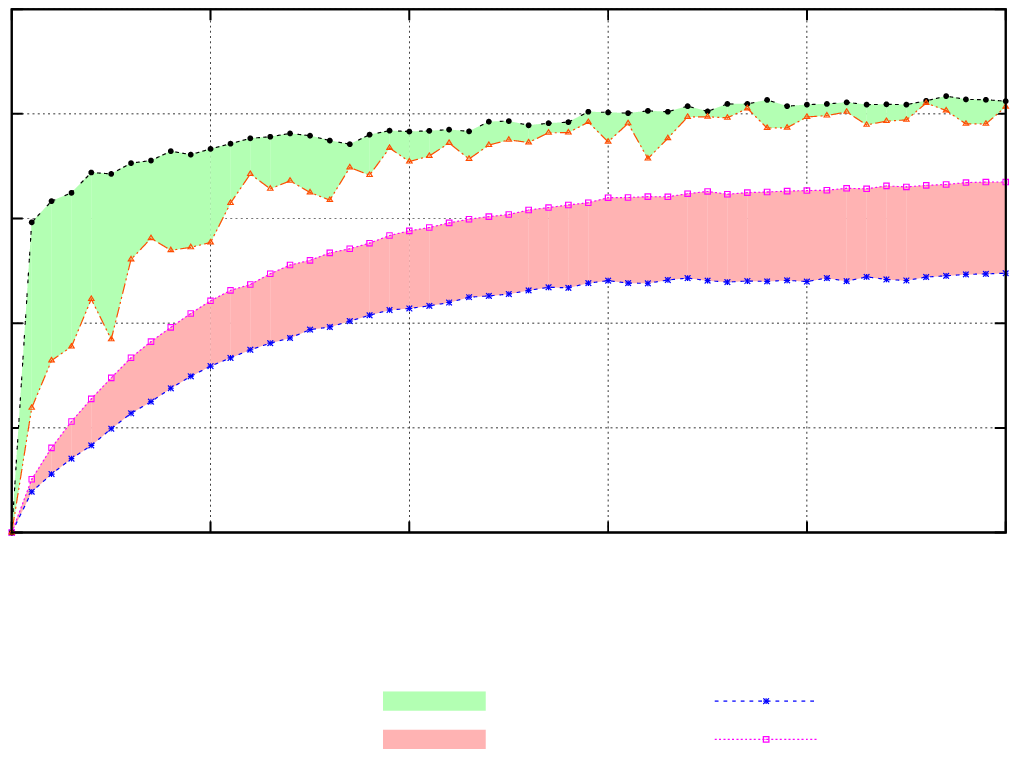}
		}
		\caption{SGP Vladislavleva-8\\crossover improvement}\label{fig:vlad8-sgp-ts3-crossover-quality}
	\end{subfigure}
	\begin{subfigure}[b]{0.5\textwidth}
		\resizebox{1\textwidth}{!}{
			\input{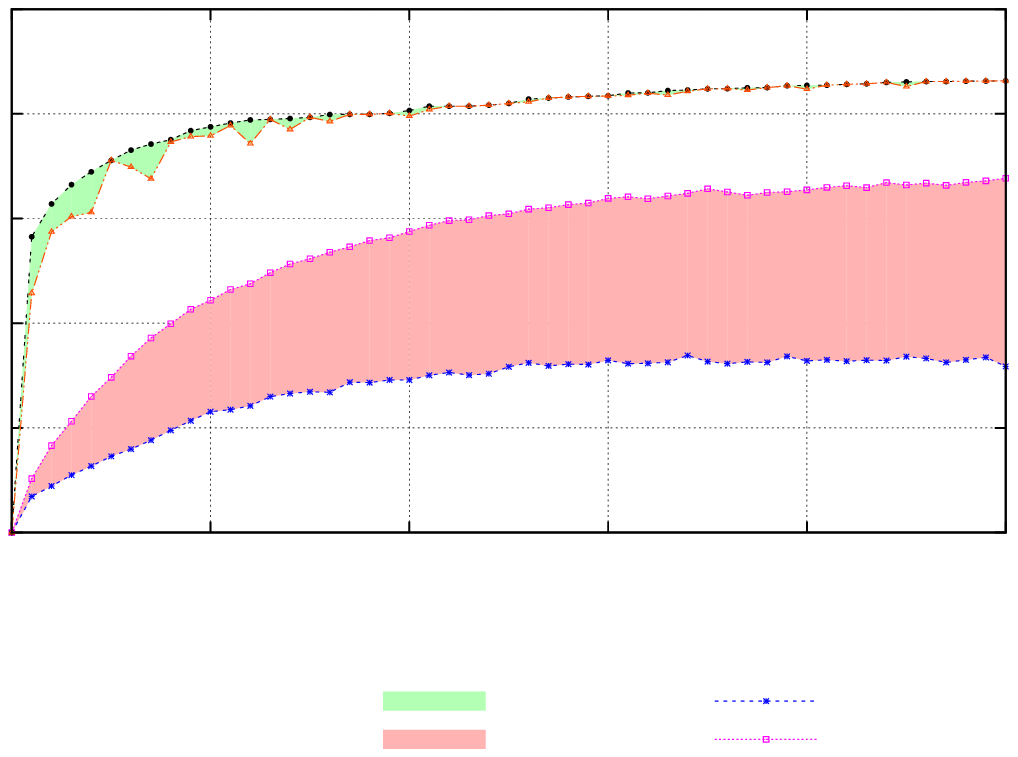}
		}
		\caption{SGP Vladislavleva-8\\mutation improvement}\label{fig:vlad8-sgp-ts3-mutation-quality}
	\end{subfigure}
	\graphicspath{{results/poly10/osgp/}}
	\begin{subfigure}[b]{0.5\textwidth}
		\resizebox{1\textwidth}{!}{
			\input{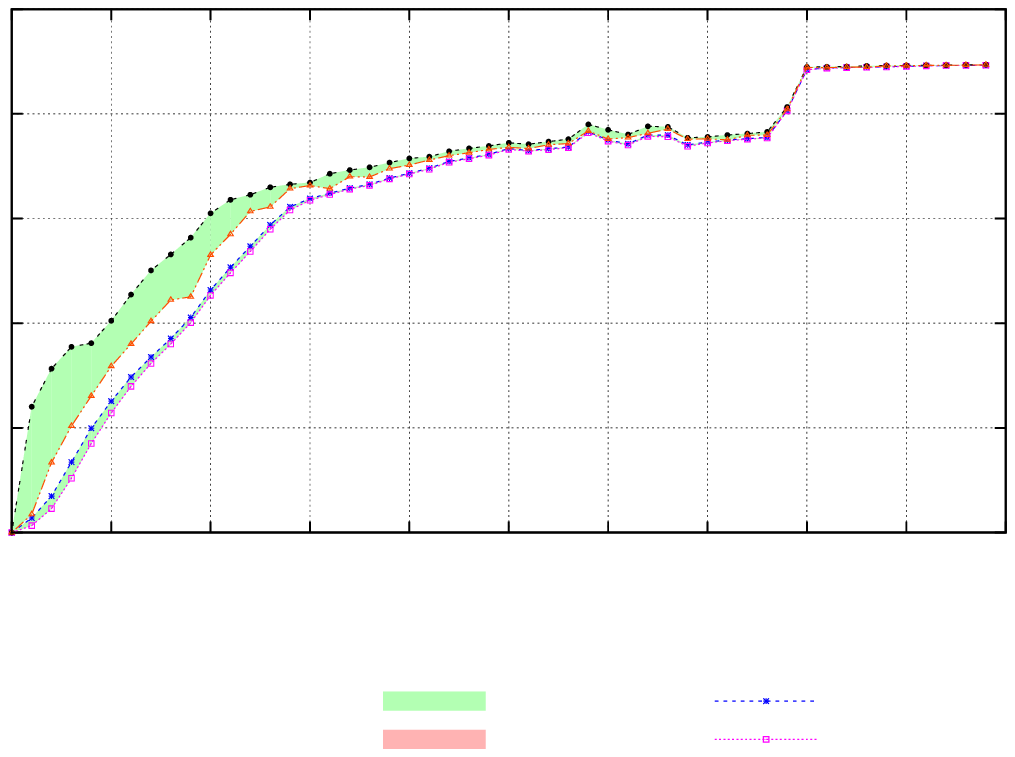}
		}
		\caption{OSGP Poly-10\\crossover improvement}\label{fig:poly10-osgp-crossover-quality}
	\end{subfigure}
	\begin{subfigure}[b]{0.5\textwidth}
		\resizebox{1\textwidth}{!}{
			\input{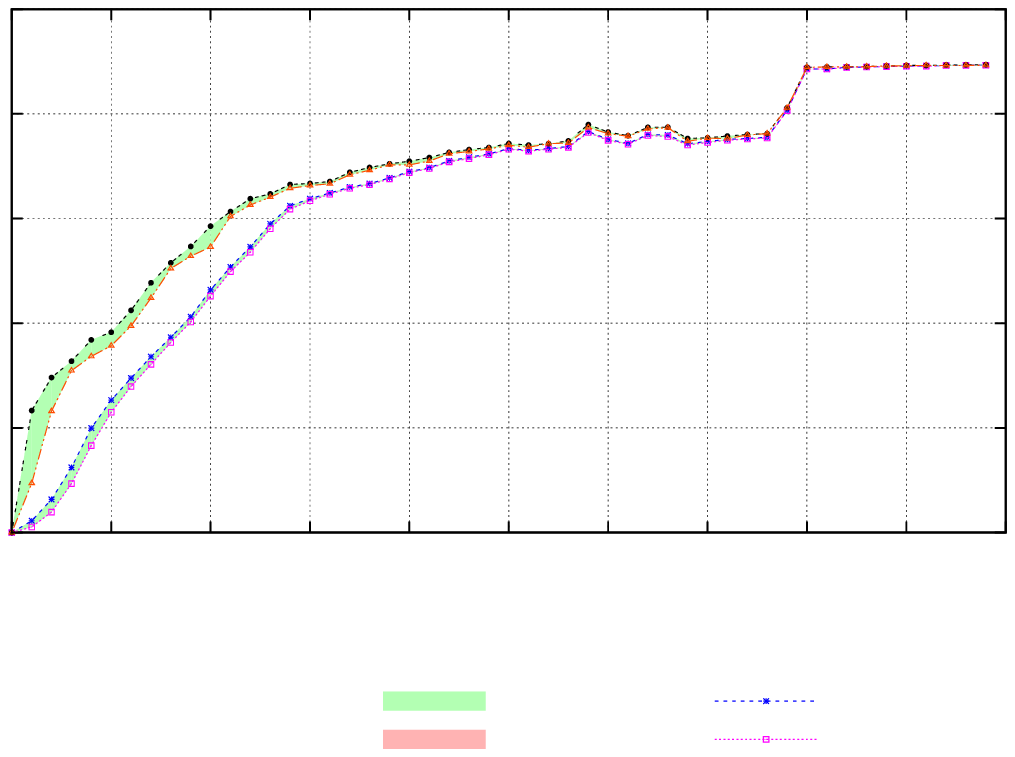}
		}
		\caption{OSGP Poly-10\\mutation improvement}\label{fig:poly10-osgp-mutation-quality}
	\end{subfigure}
	\caption{SGP Vladislavleva-8 and Poly-10 best (above) and average (below) operator improvement}\label{fig:sgp-operator-improvement}
\end{figure}

In the case of SGP, we see in Figure \ref{fig:sgp-operator-improvement} that the genetic operators produce negative improvement on average, meaning that in most cases the fitness of the child is worse than the fitness of the parent. The light-colored curves filled with green in Figure \ref{fig:sgp-operator-improvement} represent the best improvement while the dark-colored once filled with red represent the average improvement.
As average fitness improvement produced by genetic operators tends to be negative, the increase in average population fitness can be attributed to the interplay between recombination operators and selection.  
OSGP operator improvement is always small but positive due to the requirement that offspring are better than their parents.

\begin{figure}[ht]
	\graphicspath{{results/vladislavleva8/sgp/}}
	\begin{subfigure}[b]{0.5\textwidth}
		\resizebox{1\textwidth}{!}{
			\input{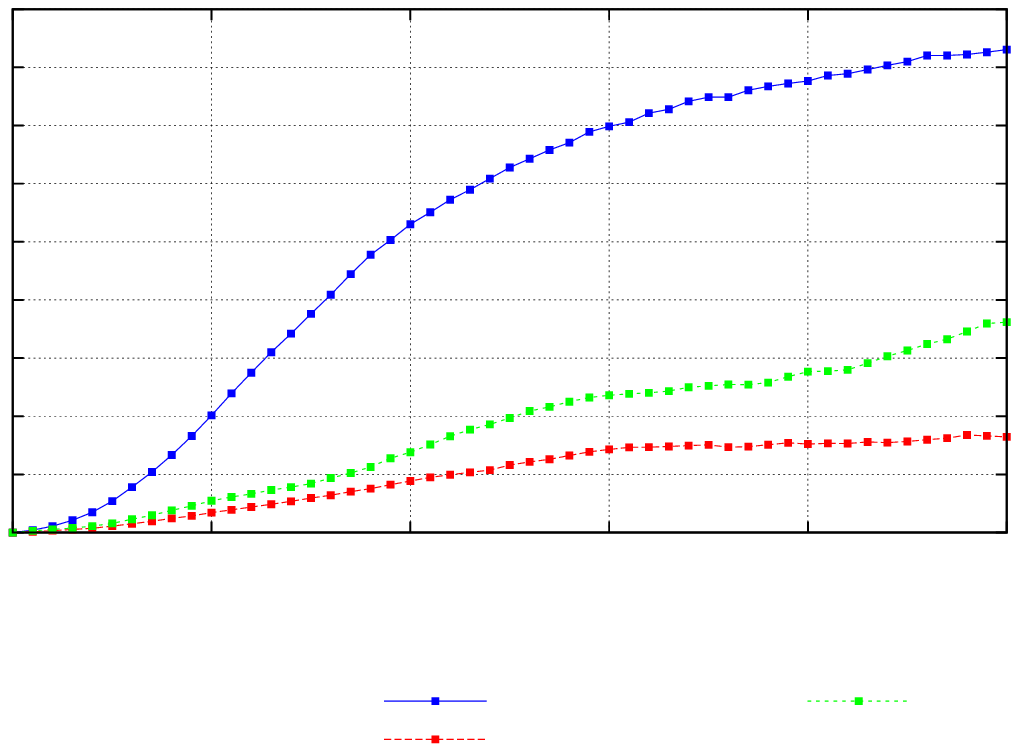}
		}
		\caption{SGP Vladislavleva-8}\label{fig:vlad8-sgp-structural-genealogical}
	\end{subfigure}
	\graphicspath{{results/poly10/osgp/}}
	\begin{subfigure}[b]{0.5\textwidth}
		\resizebox{1\textwidth}{!}{
			\input{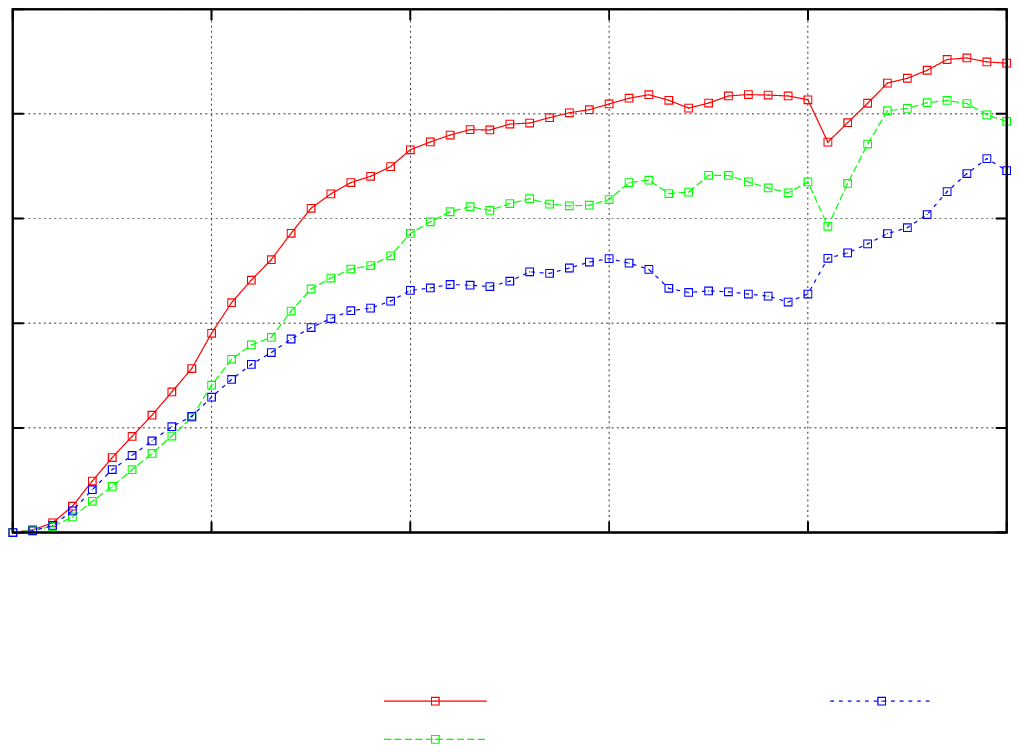}
		}
		\caption{OSGP Poly-10}\label{fig:poly10-osgp-structural-genealogical}
	\end{subfigure}
	\graphicspath{{results/vladislavleva8/sgp/}}
	\begin{subfigure}[b]{0.5\textwidth}
		\resizebox{1\textwidth}{!}{
			\input{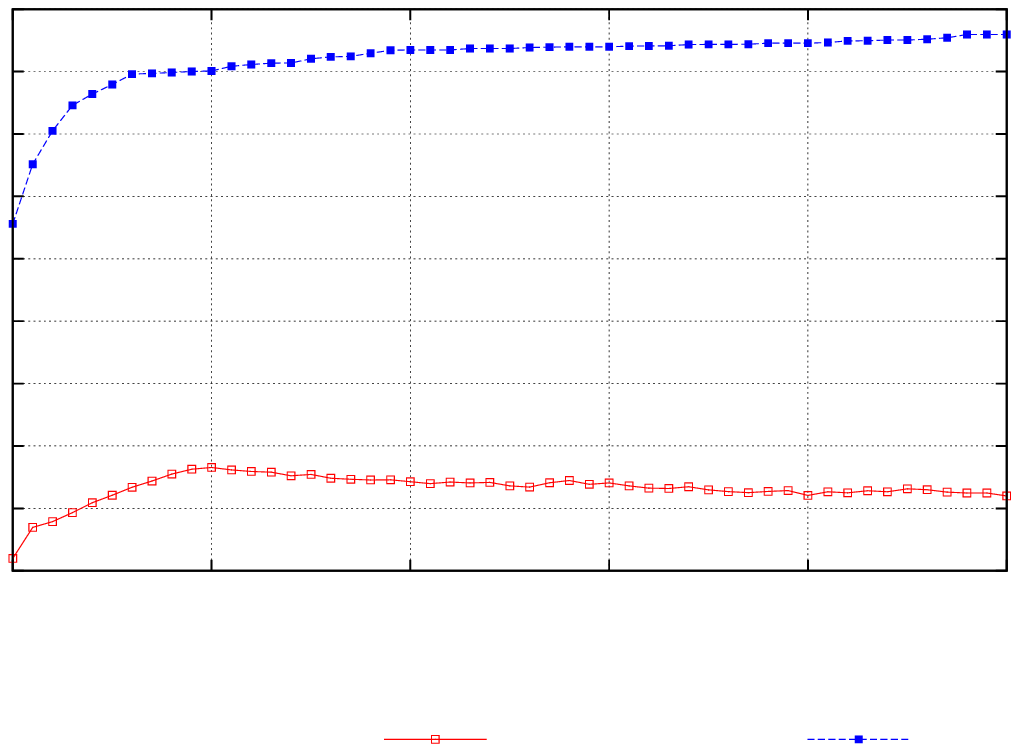}
		}
		\caption{\centering SGP Vladislavleva-8 Qualities and phenotype similarity}\label{fig:vlad8-sgp-semantic-quality}
	\end{subfigure}
	\graphicspath{{results/poly10/osgp/}}
	\begin{subfigure}[b]{0.5\textwidth}
		\resizebox{1\textwidth}{!}{
			\input{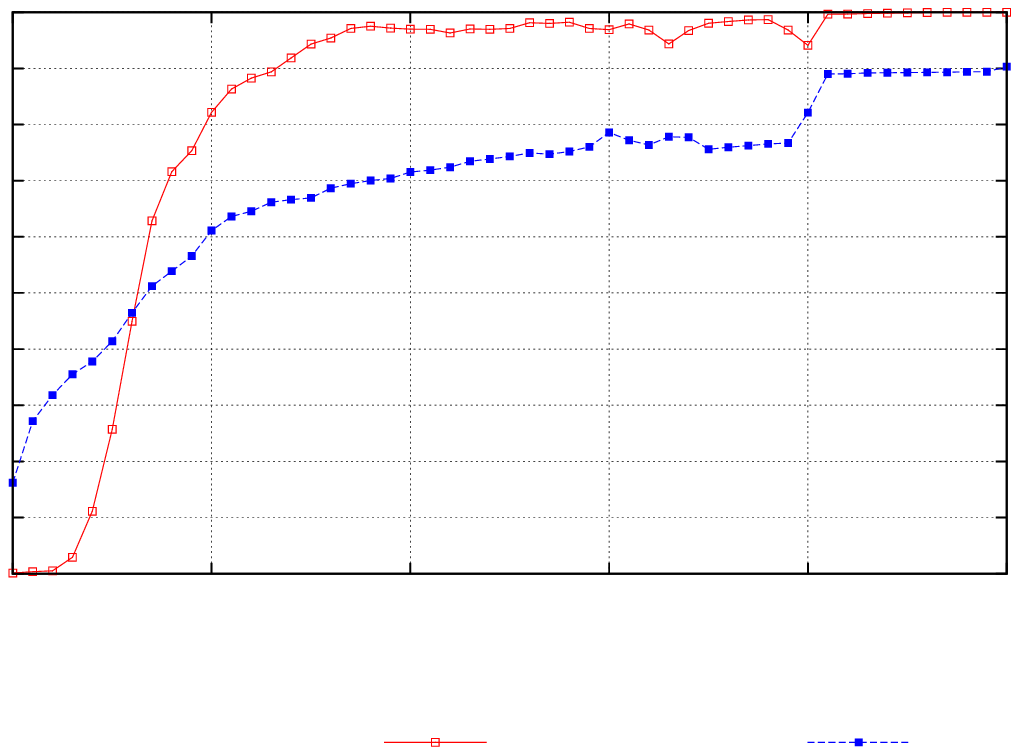}
		}
		\caption{\centering OSGP Poly-10 Qualities and phenotype similarity}\label{fig:poly10-osgp-semantic-quality}
	\end{subfigure}
	\caption{Relationship between root lineage/trace graph overlap and genotype similarity}\label{fig:sgp-structural-genealogical}
\end{figure}

The ability to produce useful genetic variation (leading to fitness improvements) is directly related to the structural diversity of the population which cannot be controlled through fitness-based selection.
Results in Figures \ref{fig:vlad8-sgp-structural-genealogical} and \ref{fig:poly10-osgp-structural-genealogical} reveal the relationship mediated by the selection mechanism between the structural similarity between two trees and the degree to which their root lineages and their trace graphs overlap. The high correlation (calculated as the Pearson $R^2$ coefficient) between the three curves corresponds intuitively to the fact that similar individuals come from similar (partially overlapping) lineages, with the important difference that trace graphs do not represent lineages in the strictest sense, as they only include those ancestors whose genes survived in the structure of the traced individual. In Figures \ref{fig:vlad8-sgp-semantic-quality} and \ref{fig:poly10-osgp-semantic-quality} we show the correlation between semantic similarity and quality of the best solution. We see that SGP does not suffer from loss of semantic diversity. With offspring selection, as children are required to outperform their parents, the semantic similarity increases rapidly to a value close to 1. 

\begin{figure}[ht]
	\centering
	\graphicspath{{results/poly10/osgp/}}
	\begin{subfigure}[b]{0.9\textwidth}
		\resizebox{1\textwidth}{!}{
			\input{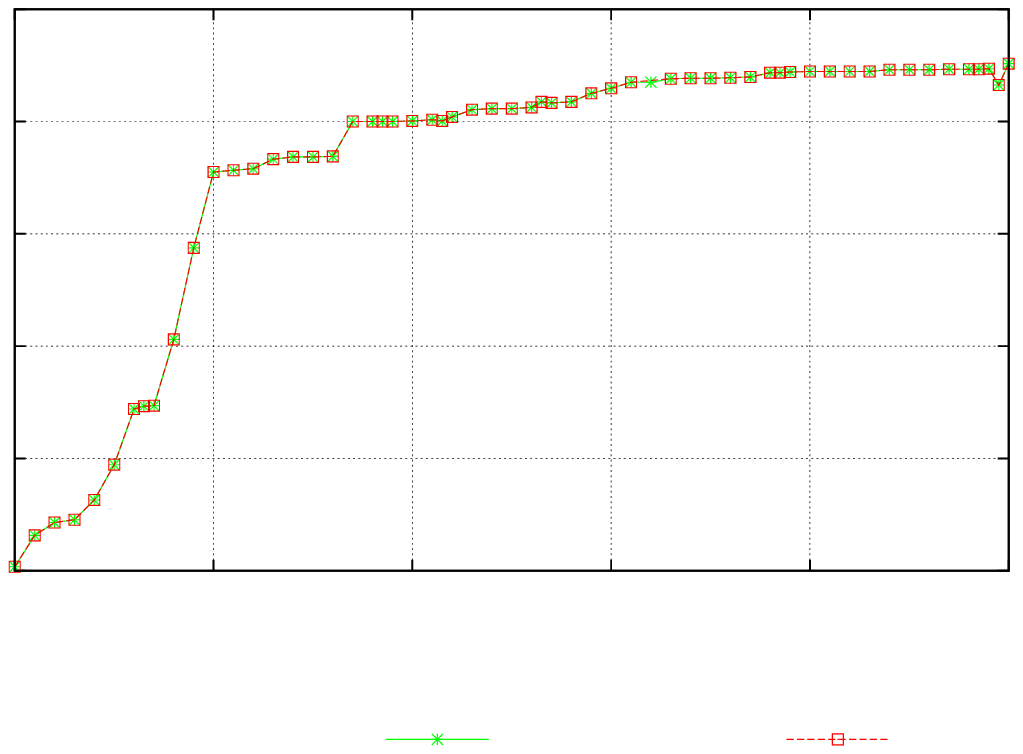}
		}
		\caption{OSGP Poly-10 best solution (the term $x_3 x_4$ was already present in the initial formula)}\label{fig:osgp-poly10-best-solution}
	\end{subfigure}
\end{figure}

Another aspect of GP search is illustrated in Figure \ref{fig:osgp-poly10-best-solution}, where we can observe the exploratory behavior of the OSGP algorithm in the beginning of the run, when the building blocks representing the terms of the formula are gradually discovered, and the exploitative behavior towards the end, when no big jumps in quality are produced, but the solution is incrementally improved through small changes of the tree constants and variable weighting factors.

Finally, the contribution ratio for SGP and OSGP was calculated at 13\% and 4\%, respectively, showing a high degree of interrelatedness between individuals which leads to low genetic operator efficiency. Fit individuals contribute multiple times, but selection pressure exceeds their variability potential. Offspring selection improves efficiency by adapting selection pressure.

\section{Conclusion and outlook}
Our results show that in most cases GP operators do not lead to fitness improvement. The tracing of the best solution indicates that a few critical operations when the algorithm is able to assemble high fitness solution elements out of preexisting, disparate genes are responsible for the performance of the entire run. A significantly small fraction (around 13\% for SGP and 4\% for OSGP) of all ancestors of the best individual have an actual contribution to its final structure.

The tracing methodology can reveal interesting and previously unexplored aspects of GP evolution regarding genetic operators and their effects on population dynamics. In contrast to other methods and techniques, our approach provides a more accurate and complete description of the evolutionary process.

\subsubsection*{Acknowledgments}
The work described in this paper was done within the COMET Project Heuristic Optimization in Production and Logistics (HOPL), \#843532 funded by the Austrian Research Promotion Agency (FFG).

\bibliographystyle{splncs}
\bibliography{bib/Burlacu,bib/gp-bibliography}

\end{document}